\documentclass[11pt, a4paper, copyright, gr]{google}

\usepackage[authoryear, sort&compress, round]{natbib}
\usepackage{amsmath,amsfonts,bm}

\def\eqref#1{equation~\ref{#1}}

\def\1{\bm{1}}

\DeclareMathAlphabet{\mathsfit}{\encodingdefault}{\sfdefault}{m}{sl}
\SetMathAlphabet{\mathsfit}{bold}{\encodingdefault}{\sfdefault}{bx}{n}

\usepackage[utf8]{inputenc} 
\usepackage[T1]{fontenc}    
\usepackage{hyperref}       
\usepackage{url}            
\usepackage{booktabs}       
\usepackage{amsfonts}       
\usepackage{nicefrac}       
\usepackage{microtype}      
\usepackage{xcolor}         

\usepackage{adjustbox}        
\usepackage{algorithm}        
\usepackage{algpseudocode}    
\usepackage{amsfonts}         
\usepackage{amsmath}          
\usepackage{amssymb}          
\usepackage{bm}               
\usepackage{booktabs}         
\usepackage{colortbl}         
\usepackage{comment}          
\usepackage{dsfont}           
\usepackage{enumitem}         
\usepackage{hyperref}         
\usepackage{listings}         
\usepackage{lscape}           
\usepackage{makecell}         
\usepackage{mathtools}        
\usepackage{multirow}         
\usepackage{multicol}         
\usepackage{nicefrac}         
\usepackage{pifont}           
\usepackage{scalerel}         
\usepackage{soul}             
\usepackage{subcaption}       
\usepackage{tabularx}         
\usepackage{tablefootnote}    
\usepackage{tcolorbox}        
\usepackage{url}              
\usepackage{wrapfig}          
\usepackage{xcolor}           
\usepackage{xspace}           
\usepackage{tabularray}
\usepackage{caption}
\usepackage{float}
\usepackage{placeins}
\usepackage{xurl}

\lstdefinestyle{pythonstyle}{
    language=Python,
    basicstyle=\ttfamily\small,
    keywordstyle=\color{blue}\bfseries,
    stringstyle=\color{orange},
    commentstyle=\color{blue},
    showstringspaces=false,
    frame=single,
    breaklines=true
}

\lstdefinestyle{base}{
  basicstyle=\ttfamily\footnotesize,
  backgroundcolor=\color{codebg},
  breaklines=true,
  breakindent=0pt,          
  breakautoindent=false,    
  columns=fullflexible,
  keepspaces=true,
  frame=single,
  rulecolor=\color{border},
  framerule=0.4pt,
  framesep=5pt,
  xleftmargin=5.4pt,
  xrightmargin=5.4pt,
  aboveskip=0pt,
}

\lstdefinestyle{md}{
  style=base,
  mathescape=false,
  literate={\$}{{\textdollar}}1,
  escapeinside={(*@}{@*)},
}

\usepackage{fontawesome5}
\usepackage[dvipsnames,table,xcdraw]{xcolor}

\newcommand*\colourcheck[1]{%
  \expandafter\newcommand\csname #1check\endcsname{\textcolor{#1}{\faCheck}}%
}
\newcommand*\colourx[1]{%
  \expandafter\newcommand\csname #1x\endcsname{\textcolor{#1}{\faTimes}}%
}

\colourcheck{ForestGreen}
\colourx{red}

\newcommand{\filehead}[2]{%
  \par\noindent
  \setlength{\fboxrule}{0.4pt}%
  \fcolorbox{border}{#1}{\parbox{\dimexpr\linewidth-2\fboxsep-2\fboxrule}{%
    \small\ttfamily\bfseries\strut #2%
  }}%
}

\newcommand{\methodcolor}[1]{\cellcolor[HTML]{FDE293}#1}
\definecolor{codebg}{HTML}{F5F5F5}
\definecolor{border}{HTML}{D0D0D0}
\definecolor{skillhead}{HTML}{FDE293}
\newcommand{\method}{WikiSkill\xspace}
\newcommand{\boldmethod}{\textbf{WikiSkill}}

\uselogo{} 

\title{WikiSkill: Compiling Agent Experience into Persistent Knowledge for Skill Evolution}

\renewcommand{\today}{2026-08-28}

\author[1]{Liyan Tang}
\author[1]{Cyrus Rashtchian}
\author[1]{Chun-Sung Ferng}
\author[1]{Andrew Tomkins}
\author[1]{Da-Cheng Juan}
\author[1,2]{Tu Vu}

\affil[1]{Google Research}
\affil[2]{Virginia Tech}
\correspondingauthor{lytang@google.com, ttvu@google.com}

\begin{abstract}
\emph{Agent skills} package specialized knowledge and workflows into reusable resources that extend AI agent capabilities. Recent work automatically discovers such skills from agent experience, which enables agents to progressively adapt through interaction. However, the insights that guide skill development typically remain scattered across optimization histories, limiting their systematic reuse across iterations. We introduce \boldmethod, a framework that co-evolves agent skills with a \emph{persistent} knowledge base (wiki). At a high level, \method separates raw execution experience, accumulated knowledge, and executable skills, while continuously consolidating experience into the wiki, which subsequent skill updates can build on. Across diverse benchmarks and models, \method consistently outperforms \emph{state-of-the-art} skill-evolution methods and improves over no-skill baselines in most model-benchmark settings. We find that skill evolution complements model scaling: larger models generally benefit more from evolved skills, while smaller models with skills can outperform substantially larger models without them. We also find that evolved skills transfer effectively across models and model families, and skills evolved by other models can outperform self-evolved skills. Finally, our ablation studies confirm that persistent knowledge accumulation in the wiki is critical for effective skill evolution. These results demonstrate the benefits of systematically accumulating and refining agent experience for developing reusable and transferable skills.
\end{abstract}

\begin{document}

\maketitle

\begin{figure}[ht!]
\centering
\includegraphics[width=0.95\textwidth]{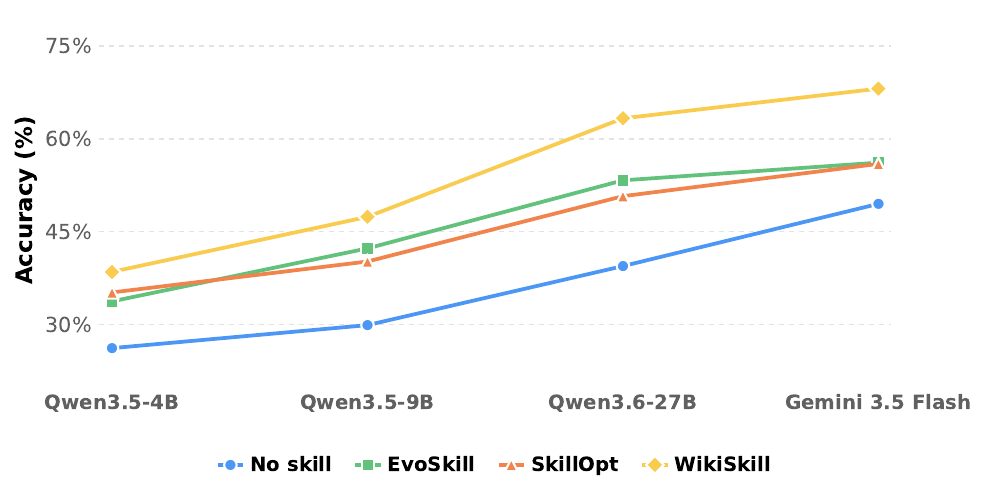}
\caption{\textbf{\method consistently improves over both the no-skill baseline and existing skill-evolution methods.} Interestingly, its advantage becomes more pronounced for stronger models. We report average accuracy across the evaluated benchmarks for each model using no skills or skills evolved by EvoSkill, SkillOpt, and WikiSkill (see Table~\ref{tab:main_results} for details).}
\label{fig:figure1}
\vspace{-2mm}
\end{figure}


\section{Introduction}
\label{sec:introduction}

General-purpose AI agents are increasingly capable of performing complex tasks across domains \citep{patwardhan2026gdpval,jackson2025aa,merrill2026terminalbench,phan2025humanity}. However, reliably accomplishing real-world tasks often requires domain-specific expertise (e.g., procedural knowledge and workflows). \emph{Agent skills}~\citep{zhang2025equipping,li2026skillsbench,chen2026skillcraftllmagentslearn, liu2026agenticskillsworkwild} provide a lightweight, open format for capturing such expertise without updating model parameters. At its core, a skill packages instructions, scripts, and other resources into a reusable filesystem-based module (i.e., an organized directory) \citep{zhang2025equipping, li2026skillsbench,anthropic2026buildingskills, xia2026metaclawjusttalk}. This design makes specialized knowledge consistent, auditable, and reusable across skill-compatible agents. It also supports \emph{progressive disclosure}~\citep{jiang2026sok}, where 
agents only load relevant content at any given time, which saves context space. More broadly, skills provide a natural mechanism for accumulating knowledge independently of model parameters.

Developing effective skills, however, remains challenging. Most agent skills are manually authored, which requires anticipating the procedural knowledge and workflows that an agent will need \citep{li2026skillsbench, liang2026skillnet, xu2026agentskillslargelanguage}. This challenge motivates recent work that iteratively develops agent skills by executing agents on training tasks, analyzing successful and failed trajectories, and refining skills based on the resulting experience~\citep{yuksekgonul2025optimizing,agrawal2026gepa,alzubi2026evoskill,ni2026trace2skill,ouyang2026skillos,yang2026skillopt}.

A key design question is how to preserve and organize what an agent learns throughout skill evolution. Prior work addresses this question in different ways. EvoSkill~\citep{alzubi2026evoskill} maintains a cumulative history of prior proposals and their evaluation outcomes; Trace2Skill~\citep{ni2026trace2skill} extracts and consolidates lessons across execution trajectories into skill updates; and SkillOpt~\citep{yang2026skillopt} uses rejected-edit feedback and epoch-wise meta guidance. However, these methods do not maintain what has been learned as a separate, evolving knowledge representation. Inspired by~\citet{karpathy2026llmwiki}'s perspective on \textbf{LLM Wiki}, which advocates compiling experience into persistent, compounding knowledge, we ask: \emph{\textbf{Can agent experience be similarly compiled into persistent knowledge to support long-term skill evolution?}} We introduce \textbf{\method}, which adds a structured knowledge layer between raw experience and executable procedures (i.e., skills). This layer allows skill development to build on increasingly well-supported and integrated knowledge across iterations, rather than on knowledge scattered across skill-evolution artifacts.

\method organizes the agent workspace into three layers: a \emph{Raw Layer} that stores immutable execution traces, a \emph{Wiki Layer} that maintains structured knowledge, and a \emph{Skill Layer} that contains evolving procedural knowledge (Figure~\ref{figure:wikiskill_diagram}). Each iteration involves four components: an \emph{Inference Agent} that executes rollouts using the current skills, a \emph{Wiki Maintainer} that consolidates traces into the wiki, a \emph{Skill Proposer} that uses the wiki and traces to propose skill updates, and a \emph{Gating and Rollback mechanism} that retains updates that improve validation performance. While skill updates can be rolled back, the wiki persists so that future updates can build on accumulated knowledge. At a high level, these components form a continual loop in which experience is consolidated into persistent knowledge that supports skill evolution.

We evaluate \method across \emph{five} benchmarks spanning mathematical reasoning (LiveMathematicanBench~\citep{he2026livemathematicianbench}), web search (SealQA~\citep{pham2026sealqa}), spreadsheet manipulation (SpreadSheetBench~\citep{ma2024spreadsheetbench}), long-context document question answering (OfficeQA~\citep{singhvi2025officeqa}), and interactive embodied tasks (ALFWorld \citep{shridhar2021alfworld}), using \emph{five} models from the Qwen~\citep{qwen2026qwen35,qwen2026qwen36}, Gemma~\citep{team2026gemma}, and Gemini~\citep{googledeepmind2026gemini35flash} families. We find that \method outperforms existing skill-evolution methods and improves over no skills in most settings. Interestingly, skill evolution complements model scaling. Within the Qwen family, \method improves average performance by 12.3\%, 17.5\%, and 23.9\% for 4B, 9B, and 27B models, respectively, \emph{with gains increasing with model scale}. At the same time, evolved skills can compensate for substantial model scale: Qwen-3.5-9B with \method outperforms Qwen-3.6-27B without skills (47.4\% vs. 39.4\%). We further find that evolved skills transfer effectively across model families and can outperform self-evolved skills. On ALFWorld, for example, Qwen-3.5-9B reaches 70.2\% with a Qwen-3.6-27B-evolved skill, compared with 63.4\% using its own skill. These results suggest that skill discovery and skill execution are distinct capabilities. Finally, our analysis shows that the persistent wiki is critical to these gains, supporting our hypothesis that accumulating and refining knowledge across iterations improves skill evolution.

In summary, our main contributions are: 
\begin{itemize}
    \item We introduce \method, a framework that co-evolves agent skills with a persistent knowledge base that continually organizes and refines knowledge from agent experience.
    \item We demonstrate across five benchmarks and five models that \method consistently outperforms existing skill-evolution methods, with ablations confirming the importance of persistent knowledge accumulation.
    \item We systematically study how evolved skills interact with model capability, showing that skill evolution complements model scaling and that evolved skills can transfer effectively across models, sometimes outperforming self-evolved skills.
\end{itemize}

Taken together, we hope that our work will spur more fundamental research on how agents can accumulate, organize, and reuse knowledge from experience.

\section{Problem Setup}
\label{sec:problem_setup}

We formalize the task of iterative skill evolution for LLM agents. Let $\mathcal{D} = \{(x_i, y_i)\}_{i=1}^N$ be a dataset of tasks, where $x_i$ denotes a task instance and $y_i$ denotes its ground-truth answer. We partition $\mathcal{D}$ into three disjoint splits: training tasks $\mathcal{D}_{\text{train}}$, validation tasks $\mathcal{D}_{\text{val}}$, and testing tasks $\mathcal{D}_{\text{test}}$.

An agent $\pi$ is an LLM-based system equipped with a set of tools $\mathcal{U}$ (e.g., a bash shell, search APIs, or file readers) and an active skill set $S = \{s_1, s_2, \dots, s_M\}$. A skill is a modular, filesystem-based directory that packages domain-specific procedural knowledge into instructions, scripts, and other resources~\citep{zhang2025equipping,li2026skillsbench,chen2026skillcraftllmagentslearn, liu2026agenticskillsworkwild}. Specifically, each skill contains a \texttt{SKILL.md} file with frontmatter metadata (a unique name and concise description) alongside full procedural instructions and applicability conditions. The skill set $S$ is initialized to empty ($\emptyset$) and developed for each dataset through the evolution process.

When executing a task $x_i$, the agent receives the task context $x_i$ and access to the available skills $S$. The agent interacts with the environment over multiple steps using its tools and skills to generate an execution trajectory $\tau_i \sim \pi(x_i; S)$. The trajectory $\tau_i = (o_1, a_1, o_2, a_2, \dots, o_T, a_T)$ consists of observations $o_t$ and actions $a_t$ (which may include calls to tools in $\mathcal{U}$). The final action $a_T$ emits a predicted answer $\hat{y}_i$. The correctness of the prediction is evaluated by a domain-specific scoring function $f(\hat{y}_i, y_i) \in [0, 1]$. For any task split $\mathcal{D}_{\text{split}} \subset \mathcal{D}$, rolling out the agent $\pi(\cdot; S)$ across all task instances in $\mathcal{D}_{\text{split}}$ yields a corresponding set of execution trajectories $\mathcal{T}_{\text{split}} = \{ \tau_i \sim \pi(x_i; S) \}_{(x_i, y_i) \in \mathcal{D}_{\text{split}}}$. The performance on a task split $\mathcal{R}(\mathcal{T}_{\text{split}})$ is the average score across all task instances $(x_i, y_i) \in \mathcal{D}_{\text{split}}$.

In \method, the system state at iteration $k$ is represented by the tuple $(S_k, W_k)$, where $S_k = \{s_1, \dots, s_M\}$ denotes the active procedural skill set and $W_k$ denotes the persistent knowledge base (Wiki). While candidate skill updates are subject to validation gating and rollback upon score degradation, the knowledge base $W_k$ persists and compounds across iterations. Starting from $(S_0, W_0) = (\emptyset, \emptyset)$, \method co-evolves the joint state $(S_k, W_k)$ across iterations $k \in \{1, \dots, K\}$, leveraging training rollouts $\mathcal{T}_{\text{train}, k}$, pattern consolidation, and validation gating based on $\mathcal{T}_{\text{val}, k}$ to maximize final test performance $\mathcal{R}(\mathcal{T}_{\text{test}})$ on unseen tasks $\mathcal{D}_{\text{test}}$.

\section{Methodology}
\label{sec:methodology}

\begin{figure*}
    \centering
\includegraphics[width=\textwidth]{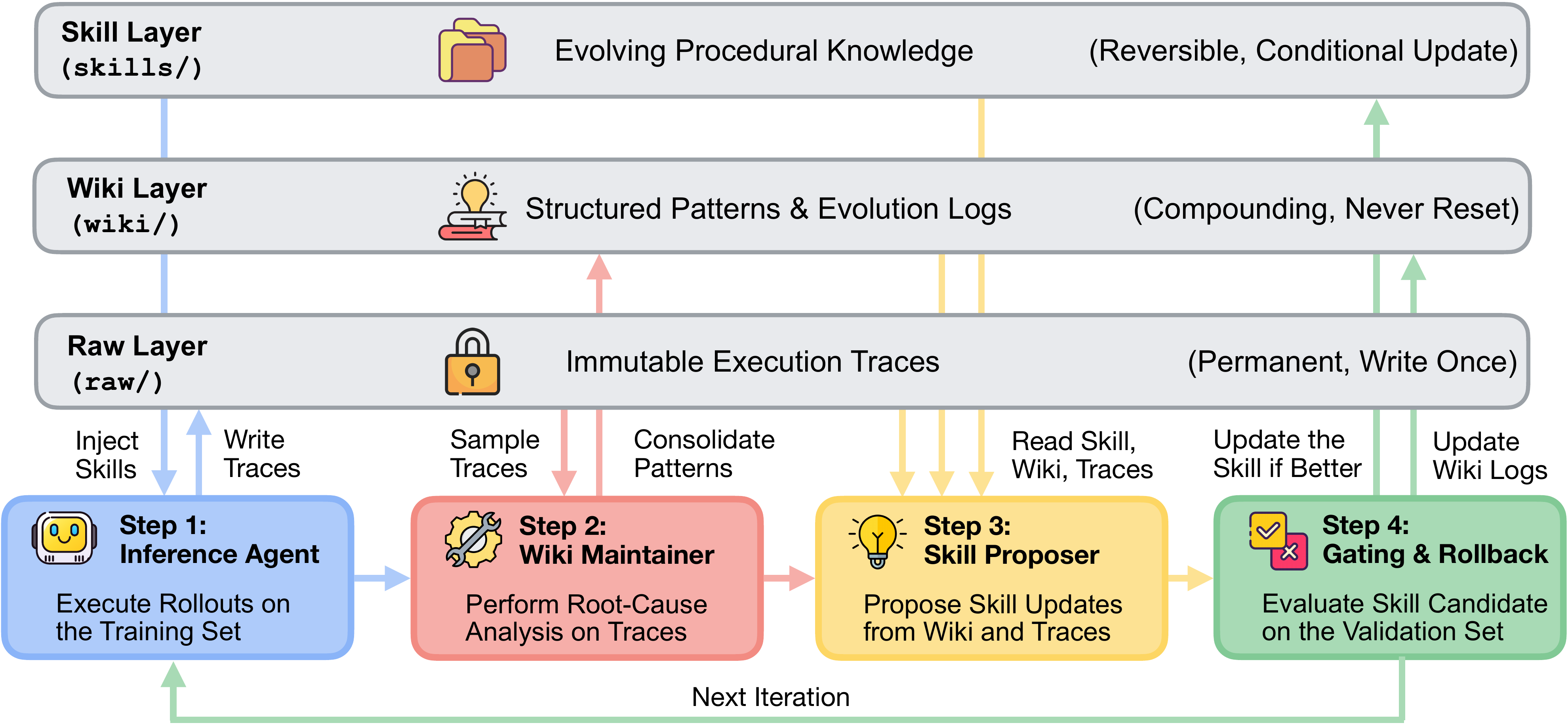}
\caption{\textbf{Overview of the WikiSkill framework.} The agent workspace is structured into three layers: immutable execution traces (Raw Layer), a persistent knowledge base that compounds across iterations (Wiki Layer), and active procedural instructions (Skills Layer). In each evolutionary loop, the Inference Agent runs rollouts (injecting active skills but restricting Wiki access), the Wiki Maintainer consolidates traces into the Wiki, and the Skill Proposer (with the ReAct mechanism) suggests updates while the Wiki is retained across all iterations.}
\label{figure:wikiskill_diagram}
\end{figure*}

We present \method, a framework that co-evolves agent skills and a persistent knowledge base (wiki). Built around a three-layer knowledge architecture (\S\ref{subsec:architecture}), \method executes an orchestrated evolutionary loop in which the agent runs rollouts, a Wiki Maintainer consolidates traces and updates the wiki, a Skill Proposer proposes skill updates, and a gating mechanism filters changes (\S\ref{subsec:agents}).

\subsection{Three-Layer Knowledge Architecture}
\label{subsec:architecture}

The \method workspace consists of three distinct layers, as shown in Figure~\ref{figure:wikiskill_diagram} and described below.

\paragraph{Raw Layer (\texttt{raw/})}
This layer stores the raw execution traces $\tau_i \in \mathcal{T}_{\text{train}, k}$ collected from training examples in each iteration. These traces capture the agent's complete step-by-step interactions, including reasoning, tool calls, tool-call outputs, and final answers. In our setup, the Wiki Maintainer and Skill Proposer agents can access these raw traces to analyze agent behavior. To preserve the raw history, this layer is immutable.

\paragraph{Wiki Layer (\texttt{wiki/})}
This layer compiles raw traces into structured, compounding knowledge and is maintained throughout skill evolution. It contains a pattern directory (\texttt{patterns/}) populated with individual markdown files that document specific failure modes or successful strategies, along with actionable workarounds. Crucially, this layer provides long-term historical awareness across optimization iterations through an evolution log (\texttt{logs.md}, updated by the Wiki Maintainer) and a skill impact tracker (\texttt{skill-impact.md}, updated programmatically by the outer-loop harness after validation gating). These records allow the Wiki Maintainer and Skill Proposer to (1) observe the complete skill acceptance history so that rejected interventions are not proposed again, (2) track what was proposed in prior iterations and whether those proposals succeeded, and (3) identify which errors recur across iterations. The wiki is not reset between iterations, but rather accumulates and compiles knowledge continuously throughout the evolution process.

\paragraph{Skills Layer (\texttt{skills/})}
This layer contains the active set of evolved skills $S$, which encode the procedural knowledge that the Inference Agent can read. Each skill directory in \method contains two files: \texttt{SKILL.md}, which contains the full content of the skill; and \texttt{PURPOSE.md}, which maps the skill back to the motivating Wiki patterns that inspired its creation or modification. A detailed example of interactions between the Skill Layer and the Wiki Layer is illustrated in Figure~\ref{figure:wiki_demo} and explained in the case study in Section~\ref{subsec:case_study}.

\subsection{Evolutionary Agents and Wiki Orchestration}
\label{subsec:agents}

The \method loop consists of four components. In each iteration, the \textbf{Inference Agent} (\S\ref{subsubsec:inference}) executes tasks using the active skills in \texttt{skills/}, producing immutable execution traces in \texttt{raw/}. During the training rollouts, the Inference Agent is restricted from accessing the Wiki Layer, as our ablation study (\S\ref{sec:ablation}) shows that allowing wiki access during training negatively affects skill development. Next, the \textbf{Wiki Maintainer} (\S\ref{subsubsec:maintenance}) analyzes these raw traces alongside the existing \texttt{wiki/} layer to diagnose failures and extract successful strategies, updating the persistent pattern catalog and evolution logs. The \textbf{Skill Proposer} (\S\ref{subsubsec:proposer}) then reviews the updated wiki and reads execution traces from the latest iteration to generate or modify candidate skills in \texttt{skills/}. Finally, a \textbf{Gating and Rollback} mechanism (\S\ref{subsubsec:gating}) evaluates the candidate skills on a validation split, accepting successful modifications or rolling back the skill set if the changes degrade performance. The entire evolution algorithm is described in Algorithm~\ref{alg:wikiskill_evolution} in Appendix~\ref{sec:method_detail}.

\subsubsection{Skill Provisioning for the Inference Agent}
\label{subsubsec:inference}

At iteration $k$, the Inference Agent $\pi$ is conditioned on the active skill set $S_{k-1}$ and executes a multi-turn trajectory using environment tools $\mathcal{U}$:
\begin{equation}
\tau_i \sim \pi(x_i; S_{k-1})
\end{equation}
In \method, the full content of active skills $S_{k-1}$ is injected directly into the Inference Agent's system prompt. Following prior work~\citep{yang2026skillopt, ni2026trace2skill}, this full-injection setting ensures that procedural instructions are immediately available during task execution, thereby eliminating skill triggering or retrieval failures as confounding variables in our study.

\subsubsection{Wiki Maintainer: Pattern Consolidation}
\label{subsubsec:maintenance}

At iteration $k$, after obtaining rollout traces $\mathcal{T}_{\text{train}, k}$ on the training split $\mathcal{D}_{\text{train}}$, we sample a subset of successful and failing execution traces $\mathcal{T}_{\text{sample}, k} \subset \mathcal{T}_{\text{train}, k}$ (see Appendix~\ref{appendix:implementation} for sampling budget and stratification criteria) to avoid context window limitations. The Wiki Maintainer agent $\mathcal{M}_{\text{WM}}$ consolidates these observations into the persistent wiki $W_{k-1}$, producing the intermediate wiki state $W'_k$:
\begin{equation}
W'_k \leftarrow \mathcal{M}_{\text{WM}}(W_{k-1}, \mathcal{T}_{\text{sample}, k})
\end{equation}
The Wiki Maintainer agent receives the full wiki context $W_{k-1}$ alongside sampled traces $\mathcal{T}_{\text{sample}, k}$. It performs root cause analysis on the failing tasks, and extracts successful strategies from the passing tasks. In each iteration, the Wiki Maintainer can create new pattern pages under \texttt{wiki/patterns/} and update existing pattern pages with new evidence or refined solutions. Updates to pattern pages are applied using incremental, patch-based editing (e.g., appending, replacing, or inserting text spans). Whenever patterns are modified, the Wiki Maintainer revises the \texttt{index.md} catalog to reflect the current state and appends a summary of the iteration's findings to the evolution log \texttt{logs.md}. There is no hard limit on the number of patterns created or updated per iteration; the Wiki Maintainer decides what updates are warranted based on the traces and the current wiki state.

\subsubsection{Wiki-Informed Skill Proposer}
\label{subsubsec:proposer}

The Proposer $\mathcal{M}_{\text{P}}$ is an LLM-based agent responsible for skill discovery and refinement. At iteration $k$, the proposer operates in a multi-turn ReAct style~\citep{yao2022react}. To avoid context window exhaustion when analyzing long execution histories, the proposer is not given a fixed set of pre-sampled traces; instead, it is initially provided with the wiki index $I(W'_k)$, the historical skill impact tracker (\texttt{skill-impact.md}), and a concise summary of all training task outcomes (pass/fail status, predictions and ground-truth answers). Operating as an autonomous agent, it actively reasons and uses environment tools (\texttt{read\_file}) to select and inspect specific pattern pages and raw execution traces $\tau_i \in \mathcal{T}_{\text{train}, k}$ on demand to diagnose root causes before synthesizing a proposal $P_k$:
\begin{equation}
P_k \leftarrow \mathcal{M}_{\text{P}}(W'_k, S_{k-1}, \mathcal{T}_{\text{train}, k})
\end{equation}
In each iteration, the Skill Proposer produces an atomic proposal $P_k$ that targets a single skill, either creating a new skill or applying an incremental, patch-based edit to the targeted existing skill.

\subsubsection{Gating and Rollback}
\label{subsubsec:gating}

Once a proposal $P_k$ is generated, it is applied to the workspace to yield a candidate skill set $S'_k = \text{Apply}(S_{k-1}, P_k)$. The system evaluates $S'_k$ on the validation split $\mathcal{D}_{\text{val}}$, obtaining validation traces $\mathcal{T}_{\text{val}, k}$ and score $\mathcal{R}(\mathcal{T}_{\text{val}, k})$. The acceptance decision is governed by:
\begin{equation}
S_k \leftarrow \begin{cases} 
S'_k & \text{if } \mathcal{R}(\mathcal{T}_{\text{val}, k}) > \mathcal{R}_{\text{best}} \\
S_{k-1} & \text{otherwise}
\end{cases}
\end{equation}
If accepted, the candidate skills are preserved as the new active skill set $S_k$, and the benchmark performance threshold $\mathcal{R}_{\text{best}}$ is updated to $\mathcal{R}(\mathcal{T}_{\text{val}, k})$. Prior to the evolution loop, $\mathcal{R}_{\text{best}}$ is initialized to the baseline validation score $\mathcal{R}(\mathcal{T}_{\text{val}, 0})$ obtained by evaluating the empty skill set $S_0$ on $\mathcal{D}_{\text{val}}$. If the validation score reaches the maximum ($\mathcal{R}_{\text{best}} = 1.0$) at any point during evolution, the evolution loop terminates early.  If rejected, the system discards the candidate skill modifications and reverts the skill set to the most recent successful configuration $S_{k-1}$. Notably, the wiki $W_k$ is never rolled back regardless of the acceptance decision; accumulated patterns and logs persist across all iterations to ensure long-term knowledge retention. Following each validation evaluation, the outer-loop orchestration harness programmatically appends an entry to \texttt{wiki/skill-impact.md} via $W_k \leftarrow \text{Update}(W'_k, P_k, \mathcal{R}(\mathcal{T}_{\text{val}, k}), a_k)$, recording the proposal metadata, target skill name, unified diff of the modification, validation score $\mathcal{R}(\mathcal{T}_{\text{val}, k})$, and final acceptance outcome $a_k \in \{\text{Accepted}, \text{Rejected}\}$. This completes the wiki state transition $W_{k-1} \to W_k$ for iteration $k$, providing an objective, ground-truth audit trail of past interventions that the Skill Proposer can consult in subsequent iterations to avoid repeating failed modifications.

\section{Experiments and Results}
\label{sec:experiments_and_results}

\subsection{Experimental Setup}

\paragraph{Datasets} We evaluate across five benchmarks spanning diverse domains: mathematical reasoning (\textbf{LiveMathematicianBench (LiveMath)}~\citep{he2026livemathematicianbench}), web search (\textbf{SealQA}~\citep{pham2026sealqa}), spreadsheet manipulation (\textbf{SpreadsheetBench (SpreadSheet)}~\citep{ma2024spreadsheetbench}), long-context document question answering \textbf{OfficeQA}~\citep{singhvi2025officeqa}), and interactive embodied tasks  (\textbf{ALFWorld}~\citep{shridhar2021alfworld}). Dataset details and statistics are provided in Appendix~\ref{appendix:datasets}.

\paragraph{Baselines} We compare \method against three representative skill-evolution baselines, including Trace2Skill~\citep{ni2026trace2skill}, EvoSkill~\citep{alzubi2026evoskill}, and SkillOpt~\citep{yang2026skillopt}, all of which share the same general loop of rolling out an agent, analyzing execution traces, proposing skill modifications, and gating changes via validation. We also evaluate each model without skills as a no-skill baseline. A detailed description and an analysis of the complexity of optimizer API calls across these frameworks are provided in Appendix~\ref{appendix:api_complexity}. We focus our comparison on dedicated skill-evolution frameworks rather than general automatic prompt optimizers (e.g., GEPA~\citep{agrawal2026gepa}), following prior work that shows specialized skill-evolution pipelines consistently outperform general prompt optimization methods~\citep{yang2026skillopt}.

\paragraph{Models} We experiment with both closed and open-weight models to evaluate \method and the baselines. For closed models, we use \textbf{Gemini-3.5-Flash}~\citep{googledeepmind2026gemini35flash}. For open-weight models, we evaluate \textbf{Qwen-3.5-4B/9B-Instruct}~\citep{qwen2026qwen35}, \textbf{Qwen-3.6-27B}~\citep{qwen2026qwen36}, and \textbf{Gemma-4-31B-It}~\citep{team2026gemma}, which we deploy using the vLLM framework~\citep{10.1145/3600006.3613165}. 

\subsection{Main Results}
\label{sec:results}

\begin{table}[t]
\centering
\small
\renewcommand{\tabcolsep}{1.8mm}
\begin{tabular}{llccccc|c}
\toprule
\textbf{Model} & \textbf{Method} & \textbf{LiveMath} & \textbf{SealQA} & \textbf{SpreadSheet} & \textbf{OfficeQA} & \textbf{ALFWorld} & \textbf{Avg.} \\
\midrule
\multirow{5}{*}{Qwen-3.5-4B} & No skill & 29.1 & 32.5 & 14.6 & 30.2 & 24.4 & 26.2 \\
 & Trace2Skill & 31.5 & \textbf{37.6} & \textbf{17.5} & 31.0 & 42.8 & 32.1 \\
 & EvoSkill & 41.7 & \textbf{37.3} & \textbf{18.6} & 29.5 & 41.5 & 33.7 \\
 & SkillOpt & \textbf{48.7} & 33.3 & 14.0 & 34.5 & 45.3 & \textbf{35.2} \\
 & \methodcolor{\textbf{WikiSkill}} & \methodcolor{\textbf{49.7}} & \methodcolor{\textbf{39.4}} & \methodcolor{\textbf{21.1}} & \methodcolor{28.5} & \methodcolor{\textbf{53.7}} & \methodcolor{\textbf{38.5}} \\
\midrule
\multirow{5}{*}{Qwen-3.5-9B} & No skill & 28.2 & 26.3 & 24.3 & 35.9 & 34.7 & 29.9 \\
 & Trace2Skill & 33.1 & 36.9 & 26.5 & 38.4 & 48.8 & 36.7 \\
 & EvoSkill & \textbf{58.1} & 34.5 & \textbf{35.4} & 34.9 & 48.5 & 42.3 \\
 & SkillOpt & 48.7 & 29.4 & 29.0 & 38.0 & 55.7 & 40.2 \\
 & \methodcolor{\textbf{WikiSkill}} & \methodcolor{\textbf{56.3}} & \methodcolor{\textbf{43.1}} & \methodcolor{\textbf{33.6}} & \methodcolor{40.5} & \methodcolor{\textbf{63.4}} & \methodcolor{\textbf{47.4}} \\
\midrule
\multirow{5}{*}{Qwen-3.6-27B} & No skill & 33.9 & 27.5 & 40.8 & 42.1 & 52.8 & 39.4 \\
 & Trace2Skill & 36.3 & 37.3 & 53.3 & \textbf{54.3} & 55.5 & 47.3 \\
 & EvoSkill & \textbf{57.3} & 32.9 & 59.5 & \textbf{52.5} & 64.2 & 53.3 \\
 & SkillOpt & 51.9 & 34.5 & 53.2 & \textbf{54.8} & 59.2 & 50.7 \\
 & \methodcolor{\textbf{WikiSkill}} & \methodcolor{\textbf{61.9}} & \methodcolor{\textbf{41.6}} & \methodcolor{\textbf{81.7}} & \methodcolor{\textbf{53.7}} & \methodcolor{\textbf{77.6}} & \methodcolor{\textbf{63.3}} \\
\midrule
\multirow{5}{*}{Gemma-4-31B} & No skill & 33.9 & 30.6 & 48.3 & 43.3 & 50.4 & 41.3 \\
 & Trace2Skill & 32.3 & 37.7 & 58.5 & 43.2 & 57.2 & 45.8 \\
 & EvoSkill & 29.8 & \textbf{38.4} & 56.4 & 39.9 & 52.6 & 43.4 \\
 & SkillOpt & 40.1 & 36.1 & \textbf{63.1} & 44.4 & 61.9 & 49.1 \\
 & \methodcolor{\textbf{WikiSkill}} & \methodcolor{\textbf{56.7}} & \methodcolor{\textbf{41.2}} & \methodcolor{\textbf{68.0}} & \methodcolor{44.2} & \methodcolor{\textbf{64.4}} & \methodcolor{\textbf{54.9}} \\
\midrule
\multirow{5}{*}{Gemini-3.5-Flash} & No skill & 33.0 & 29.4 & 50.5 & 48.6 & 85.9 & 49.5 \\
 & Trace2Skill & 41.9 & \textbf{44.3} & 56.0 & 50.0 & 85.9 & 55.6 \\
 & EvoSkill & 44.6 & \textbf{43.6} & 55.4 & 51.2 & 85.9 & 56.1 \\
 & SkillOpt & 49.7 & 28.2 & 66.1 & 49.8 & 85.9 & 55.9 \\
 & \methodcolor{\textbf{WikiSkill}} & \methodcolor{\textbf{72.6}} & \methodcolor{\textbf{44.7}} & \methodcolor{\textbf{76.6}} & \methodcolor{\textbf{60.7}} & \methodcolor{85.9} & \methodcolor{\textbf{68.1}} \\
\bottomrule
\end{tabular}
\caption{\textbf{Method comparison across inference models and test sets}. Each horizontal block evaluates a specific inference model without skills (No skill) and with skills developed by different skill-evolution methods. To ensure a fair comparison, \emph{all skill-evolution methods start with an empty skill set, and evolved skills are injected into the Inference Agent's prompt at inference time}. All reported scores are the average test performance across three independent runs of the full evolution process. Our method (\textbf{WikiSkill}) is highlighted. Bold indicates the best performance for each dataset; multiple bold results indicate methods that are not significantly different from the best under a paired bootstrap test with 1,000 iterations ($p < 0.05$).
}
\label{tab:main_results}
\vspace{-5pt}
\end{table}

We evaluate \method across models and tasks and study whether evolved skills transfer across models. Table~\ref{tab:main_results} presents the main skill-evolution results across models and tasks, including how the benefits of skill evolution vary with model scale, while Table~\ref{tab:cross_model_transfer} presents the cross-model skill transfer results. We analyze these results in detail below.

To account for variability, we repeat the full evolution process across three independent runs for each method, and all reported scores represent the average test performance across the three resulting evolved skill sets. Statistical significance of performance differences is evaluated using paired bootstrap testing at $p < 0.05$ (Appendix~\ref{appendix:implementation}). Note that for Gemini-3.5-Flash on ALFWorld, all evolution methods yield the same performance (85.9\%) as the no-skill baseline because Gemini-3.5-Flash achieves a 100\% score on the validation split ($\mathcal{D}_{\text{val}}$) before skill evolution. This also explains why Gemini-3.5-Flash is marked with `$-$' as a skill source on ALFWorld in the cross-model transfer evaluation (Table~\ref{tab:cross_model_transfer}).

\subsubsection{Skill Evolution Across Models and Tasks}
\vspace{-5pt}
\paragraph{\method yields consistent improvements across models and datasets} As shown in Table~\ref{tab:main_results}, \method achieves the highest average performance across all five models. Compared with the strongest competing skill-evolution method for each model, \method improves average performance by 3.3, 5.1, 10.0, 5.8, and 12.0 points for Qwen-3.5-4B, Qwen-3.5-9B, Qwen-3.6-27B, Gemma-4-31B, and Gemini-3.5-Flash, respectively. These improvements are consistent across settings: \method improves over the no-skill baseline in most model-dataset pairs and matches or exceeds the strongest competing method across all models in the average performance across 5 datasets. The improvements also span diverse domains and can be substantial. For example, \method improves Gemini-3.5-Flash from 33.0\% to 72.6\% on LiveMath and from 50.5\% to 76.6\% on SpreadSheet, while improving Qwen-3.6-27B from 52.8\% to 77.6\% on ALFWorld. In contrast, existing skill-evolution methods are less consistent. For example, EvoSkill improves Qwen-9B substantially on LiveMath (28.2\% $\rightarrow$ 58.1\%) but degrades Gemma-4-31B on the same benchmark (33.9 \% $\rightarrow$ 29.8\%), while SkillOpt degrades Gemini-3.5-Flash on SealQA (29.4 \% $\rightarrow$ 28.2\%). These results show that WikiSkill produces both stronger and more reliable improvements across settings.

\vspace{-5pt}
\paragraph{The benefits of skill evolution increase with model capability and complement model scaling} Within the Qwen family, the average improvement from \method increases with model scale, from +12.3 points for Qwen-3.5-4B to +17.5 points for Qwen-3.5-9B and +23.9 points for Qwen-3.6-27B. This trend is particularly pronounced on SpreadSheet, where \method improves the three models by +6.5, +9.3, and +40.9 points, respectively, showing that the benefits of skill evolution can increase substantially with model scale. At the same time, evolved skills can compensate for substantial differences in model scale: Qwen-3.5-9B with \method reaches 47.4\% average accuracy, outperforming Qwen-3.6-27B without skills at 39.4\%, while Qwen-3.5-4B with \method reaches 38.5\%. Our results suggest that model capability and evolved procedural knowledge provide complementary sources of performance: stronger models can derive greater value from skill evolution by developing and executing more effective skills, while effective skills can allow smaller models to outperform substantially larger models that do not use skills.

\vspace{-5pt}
\paragraph{The benefits of skill evolution also vary substantially across datasets} Our results suggest that some datasets are more amenable to skill evolution than others. For Qwen-3.6-27B, \method improves performance by 11.6 points on OfficeQA and 14.1 points on SealQA, compared with 24.8 points on ALFWorld, 28.0 points on LiveMath, and 40.9 points on SpreadSheet. Similar differences appear across other models. LiveMath consistently benefits from skill evolution, with gains ranging from 20.6 to 39.6 points across all five models, while ALFWorld yields gains of 14.0 to 29.3 points across the four models for which WikiSkill evolves skills (excluding Gemini-3.5-Flash due to early stopping). In contrast, OfficeQA presents unique challenges due to its long-context document-retrieval requirements. Larger models effectively leverage evolved search workflows to navigate lengthy documents (e.g., +11.6 points for Qwen-3.6-27B and +12.1 points for Gemini-3.5-Flash), whereas Qwen-3.5-4B struggles to execute these multi-step search workflows across long contexts and reverts to its default reading behavior, resulting in slight degradation.

\begin{table}[t]
\centering
\small
\renewcommand{\tabcolsep}{1.8mm}
\begin{tabular}{llccccc}
\toprule
\textbf{Model} & \textbf{Skill Source} & \textbf{LiveMath} & \textbf{SealQA} & \textbf{SpreadSheet} & \textbf{OfficeQA} & \textbf{ALFWorld} \\
\midrule
\multirow{4}{*}{Qwen-3.5-4B} & None & 29.1 & 32.5 & 14.6 & 30.2 & 24.4 \\
 & \methodcolor{Qwen-3.5-4B} & \methodcolor{49.7} & \methodcolor{\textbf{39.4}} & \methodcolor{21.1} & \methodcolor{28.5} & \methodcolor{\textbf{53.7}} \\
 & Qwen-3.6-27B & 59.7 & \textbf{38.8} & \textbf{33.0} & 25.4 & \textbf{57.0}\\
 & Gemini-3.5-Flash & \textbf{62.6} & \textbf{37.3} & 23.0 & 32.2 & - \\
\midrule
\multirow{5}{*}{Qwen-3.5-9B} & None & 28.2 & 26.3 & 24.3 & 35.9 & 34.7 \\
 & Qwen-3.5-4B & \textbf{61.0} & \textbf{40.4} & 25.0 & 40.3 & \textbf{69.2} \\
 & \methodcolor{Qwen-3.5-9B} & \methodcolor{56.3} & \methodcolor{\textbf{43.1}} & \methodcolor{33.6} & \methodcolor{40.5} & \methodcolor{63.4} \\
 & Qwen-3.6-27B & 59.1 & \textbf{40.4} & \textbf{50.5} & 39.9 & \textbf{70.2} \\
 & Gemini-3.5-Flash & 53.0 & {\textbf{39.6}} & \textbf{48.8} & 40.5 & - \\
\midrule
\multirow{4}{*}{Qwen-3.6-27B} & None & 33.9 & 27.5 & 40.8 & 42.1 & 52.8 \\
 & Qwen-3.5-4B & \textbf{62.6} & 41.6 & 40.6 & \textbf{52.9} & 72.1 \\
 & \methodcolor{Qwen-3.6-27B} & \methodcolor{\textbf{61.9}} & \methodcolor{41.6} & \methodcolor{\textbf{81.7}} & \methodcolor{\textbf{53.7}} & \methodcolor{\textbf{77.6}} \\
 & Gemini-3.5-Flash & \textbf{65.1} & \textbf{51.0} & 76.0 & \textbf{52.5} & - \\
\midrule
\multirow{5}{*}{Gemma-4-31B} & None & 33.9 & 30.6 & 48.3 & 43.3 & 50.4 \\
 & Qwen-3.5-4B & \textbf{73.1} & \textbf{38.8} & 37.1 & 42.1 & \textbf{66.9}\\
 & Qwen-3.6-27B & \textbf{73.7} & 37.7 & \textbf{72.0} & 44.2 & \textbf{66.9} \\
 & \methodcolor{Gemma-4-31B} & \methodcolor{56.7} & \methodcolor{\textbf{41.2}} & \methodcolor{\textbf{68.0}} & \methodcolor{44.2} & \methodcolor{64.4} \\
 & Gemini-3.5-Flash & 61.8 & 37.7 & \textbf{68.8} & 43.4 & - \\
\midrule
\multirow{4}{*}{Gemini-3.5-Flash} & None & 33.0 & 29.4 & 50.5 & 48.6 & 85.9 \\
 & Qwen-3.5-4B & 67.5 & 40.0 & 18.1 & 48.5 & 87.3\\
 & Qwen-3.6-27B & \textbf{73.9} & \textbf{43.5} & 63.4 & 47.7 & 86.8\\
 & \methodcolor{Gemini-3.5-Flash} & \methodcolor{\textbf{72.6}} & \methodcolor{\textbf{44.7}} & \methodcolor{\textbf{76.6}} & \methodcolor{\textbf{60.7}} & \methodcolor{-} \\
\bottomrule
\end{tabular}
\caption{
\textbf{Cross-model skill transfer results.} We evaluate inference models using no skills (None) and skills evolved by \method with Qwen-3.5-4B, Qwen-3.6-27B, and Gemini-3.5-Flash as source models. Skills are injected into the Inference Agent's system prompt at inference time. Highlighted rows indicate self-evolved skills, where the inference model and skill source are the same. The highest performance per benchmark within each model block is bolded. `$-$' indicates that the source model reached 100\% validation performance before skill evolution, so no skill was evolved.
}
\label{tab:cross_model_transfer}
\vspace{-5pt}
\end{table}

\subsubsection{Cross-Model Skill Transfer with \method}
\label{subsec:cross_model_transfer}

\vspace{-5pt}
\paragraph{Evolved skills transfer effectively across models, and transferred skills can outperform self-evolved skills} Table~\ref{tab:cross_model_transfer} evaluates how skills evolved by WikiSkill transfer across inference models when developed using different source models. Transferred skills frequently outperform both the no-skill baseline and self-evolved skills. For example, Qwen-3.6-27B skills improve Qwen-3.5-9B to 50.5\% on SpreadSheet, compared with 24.3\% without skills and 33.6\% with self-evolved skills, and improve Gemma-4-31B to 73.7\% on LiveMath, compared with 33.9\% and 56.7\%, respectively. Notably, effective transfer also occurs from smaller to larger models: Qwen-3.5-4B skills improve Gemma-4-31B to 73.1\% on LiveMath and 66.9\% on ALFWorld. Our results indicate that stronger source models do not necessarily produce better skills and that procedural knowledge developed by one model's experience can transfer across model scales and families.

\paragraph{The transferability of evolved skills depends on whether they capture general procedures or model-specific workarounds} Our results in Table~\ref{tab:cross_model_transfer} suggest that \method can produce both general procedural knowledge that transfers across models and model-specific strategies that can cause negative transfer. LiveMath skills transfer particularly well across models: Qwen-3.5-4B and Qwen-3.6-27B skills improve Gemini-3.5-Flash from 33.0\% to 67.5\% and 73.9\%, respectively. In contrast, SpreadSheet exhibits strong source-target interactions. Qwen-3.5-4B skills reduce Gemini-3.5-Flash performance from 50.5\% to 18.1\%, while Qwen-3.6-27B skills improve it to 63.4\%. Our error analysis identifies two factors behind this negative transfer. First, Qwen-3.5-4B skills encode low-level workarounds, such as single-line Python commands and string-conversion rules, which help the smaller model avoid execution failures but constrain stronger models such as Gemini-3.5-Flash from using comprehensive end-to-end scripts. Second, fragmented diagnostic procedures introduce redundant tool calls that can exhaust Gemini-3.5-Flash's interaction budget before task completion.

\paragraph{The utility of transferred skills also depends on the inference model's ability to execute them} We now turn toward how different inference models use skills developed by the same source model. Within the Qwen family, stronger models can derive greater value from the same procedural knowledge. For example, Qwen-3.6-27B SpreadSheet skills improve Qwen-3.5-4B, Qwen-3.5-9B, and Qwen-3.6-27B over their no-skill baselines by 18.4\%, 26.2\%, and 40.9\%, respectively. OfficeQA provides a case where a model develops skills that are more useful to another model than to itself: Qwen-3.5-4B skills decrease its own performance from 30.2\% to 28.5\%, but improve Qwen-3.6-27B from 42.1\% to 52.9\%. Our trajectory analysis suggests that in long-context settings, smaller models can become distracted by lengthy document contexts and fail to follow detailed multi-step search instructions, instead reverting to their default document-reading behavior. Stronger models, in contrast, more reliably execute the structured navigation procedures specified by the skill across long contexts. Taken together, these results distinguish two capabilities that self-evolution normally conflates: discovering useful procedural knowledge from experience and effectively executing that knowledge at inference time.

\vspace{-5pt}
\section{Analysis and Discussion}

\subsection{Role of Persistent Knowledge in Skill Evolution}
\label{sec:ablation}

To understand where persistent knowledge contributes to skill evolution, we ablate wiki access for the two components that can use it during evolution (Table~\ref{tab:ablation}). Specifically, using Gemini-3.5-Flash, we independently vary wiki access for the Inference Agent during training rollouts and the Skill Proposer during skill development, which results in four configurations. When the Skill Proposer has no wiki access, we also remove the Wiki Maintainer, eliminating persistent knowledge accumulation across iterations. Our default \method configuration gives wiki access to the Skill Proposer but not the Inference Agent.

\begin{table}[t]
\centering
\small
\begin{tabular}{cclcccc|c}
\toprule
\multicolumn{2}{c}{\textbf{WikiSkill Components}} & & \multicolumn{5}{c}{\textbf{Benchmarks}} \\
\cmidrule(r){1-2} \cmidrule(l){4-8}
\textbf{Inference Agent} & \textbf{Skill Proposer} & & \textbf{LiveMath} & \textbf{SealQA} & \textbf{SpreadSheet} & \textbf{OfficeQA} & \textbf{Avg.} \\
\textbf{Wiki Access} & \textbf{Wiki Access} & & & & & & \\
\midrule
\multicolumn{2}{c}{No skill} & & 33.0 & 29.4 & 50.5 & 48.6 & 40.4 \\
\ForestGreencheck & \redx & & 43.8 & 42.0 & 44.4 & 51.0 & 45.3 \\
\redx & \redx & & 51.3 & 38.4 & 49.9 & 55.2 & 48.7 \\
\ForestGreencheck & \ForestGreencheck & & 64.8 & 42.8 & \textbf{80.2} & 55.6 & 60.9 \\
\textbf{\redx} & \textbf{\ForestGreencheck} & & \textbf{72.6} & \textbf{44.7} & 76.6 & \textbf{60.7} & \textbf{63.7} \\
\bottomrule
\end{tabular}
\caption{
\textbf{Ablation study on \method using Gemini-3.5-Flash.} We evaluate performance across benchmarks under four configurations that vary whether the Inference Agent and Skill Proposer have wiki access during skill evolution. When the Skill Proposer has no Wiki access, we also remove the Wiki Maintainer, eliminating persistent knowledge accumulation across iterations. The bottom row represents our default \method configuration.
}
\label{tab:ablation}
\vspace{-5pt}
\end{table}

\paragraph{Persistent wiki knowledge dramatically improves skill evolution} 
As shown in Table 3, when wiki access for the Inference Agent is disabled, providing the Skill Proposer with access to the persistent wiki increases average benchmark performance from 48.7\% to 63.7\% (+15.0\%), with substantial gains on LiveMath (51.3\% to 72.6\%) and SpreadsheetBench (49.9\% to 76.6\%). Without persistent knowledge accumulated across iterations, the Skill Proposer struggles to resolve intricate failure modes.

\paragraph{Wiki access for the Inference Agent during evolution degrades final skill quality} 
When the Skill Proposer has access to the persistent wiki, providing the Inference Agent with wiki access during training rollouts reduces average benchmark performance from 63.7\% to 60.9\%, with a substantial drop on LiveMath from 72.6\% to 64.8\%. We hypothesize that when the Inference Agent has access to both skills and the wiki during training rollouts, some task-solving knowledge may be obtained directly from the wiki rather than the skills, which can make the resulting trajectories less informative for skill development.

\subsection{Qualitative Analysis: Skill and Wiki Dynamics}
\label{subsec:qualitative_analysis}

To better understand how \method evolves knowledge and skills across models and datasets, we analyze the wiki patterns accumulated and skills produced during evolution (Table~\ref{tab:skill_pattern_summary}) and when successful skill updates are accepted across iterations (Appendix Table~\ref{tab:temporal_timeline}).

\paragraph{WikiSkill continuously accumulates wiki patterns while producing concise skills} 
Table~\ref{tab:skill_pattern_summary} summarizes the creation and editing of skills and wiki patterns across models and benchmarks, along with their average lengths. Across models, Qwen models produce longer procedural skills (118.9-128.6 lines), whereas Gemma-4-31B and Gemini-3.5-Flash produce more compact skills (45.1 and 81.2 lines, respectively). Wiki pattern accumulation also varies across models, with 6.3-8.9 patterns created and 7.0–18.4 edits on average. Across benchmarks, SpreadSheet produces the longest skills (142.5 lines) and most wiki patterns (9.8), whereas LiveMath produces the shortest skills (84.6 lines) and fewest wiki patterns (4.4). Overall, these results show that both skill structure and wiki accumulation vary across models and datasets.

\begin{figure*}[t]
    \centering
\includegraphics[width=\textwidth]{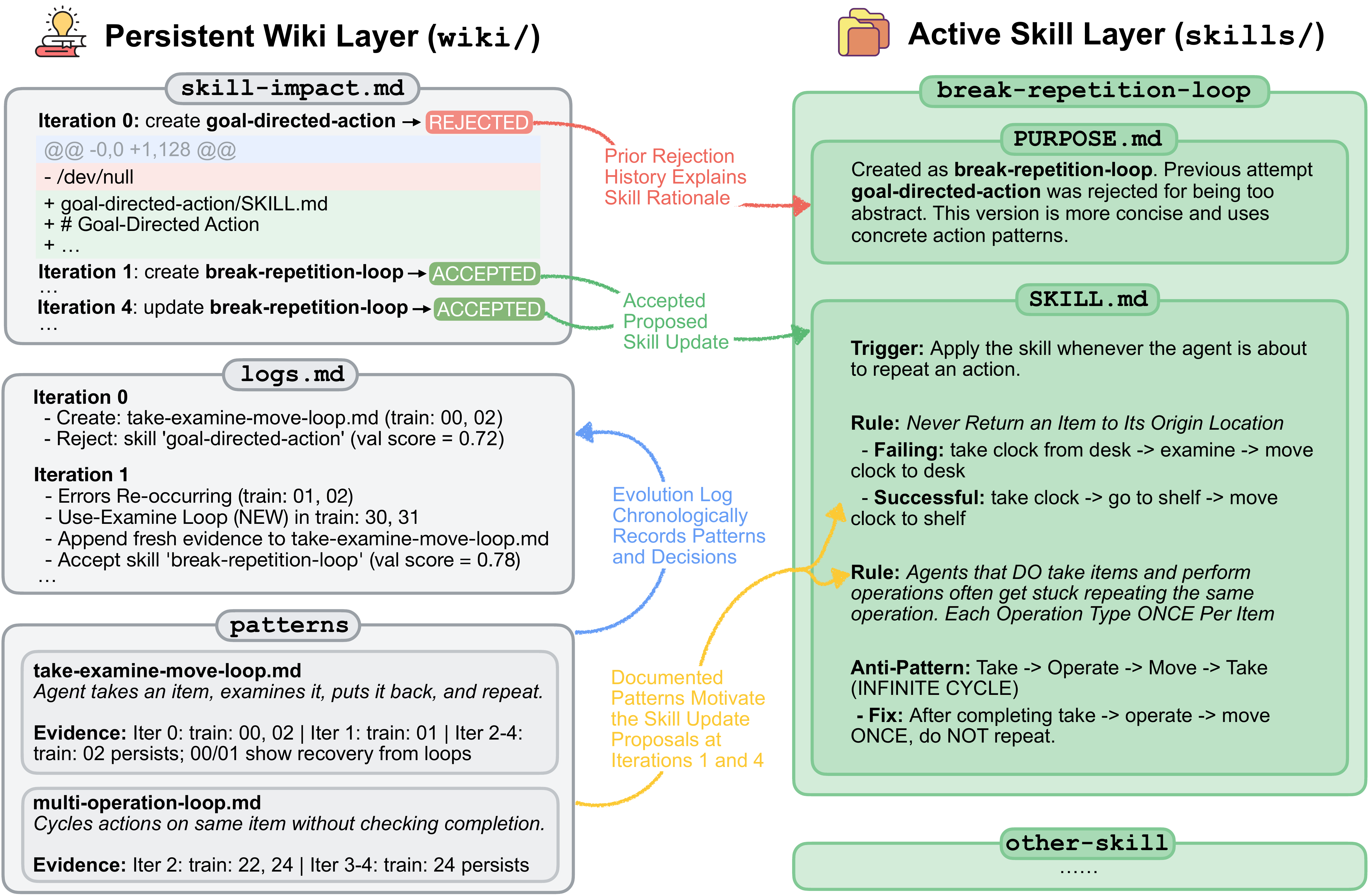}
\caption{Case study of Wiki-guided skill evolution on ALFWorld (Qwen-3.6-27B). The persistent Wiki Layer compiles cross-iteration patterns, an audit trail of past proposal diffs and acceptance decisions, and chronological history. Informed by the rejection of the skill proposal at Iteration 0, the proposer synthesizes the accepted skill update at Iteration 1, and later refines it with new pattern evidence. File contents are simplified for clarity.}
\label{figure:wiki_demo}
\end{figure*}

\paragraph{Skill refinement continues throughout the evolution process} Appendix Table~\ref{tab:temporal_timeline} groups accepted skill updates into early (Iterations 0–1), middle (Iterations 2–4), and late (Iterations 5–7) stages. Across models, the initial stage accounts for 39\%-52\% of accepted updates, with substantial fractions continuing into the middle and late stages. A similar pattern holds across benchmarks, where 39\%-58\% of accepted updates occur during the initial stage. Continued refinement is particularly pronounced on SealQA, where 33\% of accepted updates occur in the middle stage and 28\% in the late stage. Combined with the ablation in Section~\ref{sec:ablation}, these results suggest that persistent knowledge accumulation supports continued skill refinement across iterations. The Wiki Layer preserves recurring errors, rejected proposals, and evolution history, which provide the Skill Proposer with accumulated context for subsequent updates. Below, we present a case study that illustrates how this accumulated knowledge informs skill evolution.

\subsection{Case Study: Anatomy of Wiki-Guided Skill Evolution}
\label{subsec:case_study}
To illustrate how the Wiki and Skill Layers interact during evolution, we trace a concrete example from Qwen-3.6-27B on ALFWorld, as shown in Figure~\ref{figure:wiki_demo} (simplified for presentation).

At Iteration 0, the Wiki Maintainer identifies a basic looping behavior (\texttt{\nolinkurl{take-examine-move-loop.md}}), while the Skill Proposer proposes \texttt{goal-directed-action}, which fails to improve performance on the validation set and is rejected. Crucially, \textit{skill-impact.md} preserves the proposal diff and rejection outcome, allowing subsequent skill updates to account for this failed attempt.

Informed by this audit trail, the Skill Proposer creates \texttt{break-repetition-loop} at Iteration 1 with a concrete action rule (\emph{Never Return an Item to Its Origin Location}), which is accepted. As new loop variants emerge across rollouts (\texttt{multi-operation-loop.md}), the Wiki Maintainer accumulates new evidence in the persistent wiki. Guided by these accumulated wiki patterns and newly stored trajectories (not shown in the figure), the Skill Proposer further refines the skill at Iteration 4 with a new rule (\emph{Each Operation Type ONCE Per Item}). This example illustrates how persistent knowledge from prior iterations informs subsequent skill refinement.

\begin{table}[t]
\centering
\small
\begin{adjustbox}{width=1.0\textwidth}
\begin{tabular}{lcccccc}
\toprule
\multirow{2}{*}{\textbf{Category}} & \multicolumn{3}{c}{\textbf{Skills}} & \multicolumn{3}{c}{\textbf{Wiki Patterns}} \\
\cmidrule(lr){2-4} \cmidrule(lr){5-7}
 & \textbf{\shortstack{Create\\(Proposed / Accepted)}} & \textbf{\shortstack{Edit\\(Proposed / Accepted)}} & \textbf{\shortstack{Avg.\\Length}} & \textbf{Create} & \textbf{Edit} & \textbf{\shortstack{Avg.\\Length}} \\
\midrule
\multicolumn{7}{l}{\textbf{\textit{By Model (All-Dataset Average)}}} \\
Qwen-3.5-4B & 3.1 / 1.6 & 4.9 / 1.3 & 126.2 & 8.8 & 18.4 & 48.2 \\
Qwen-3.5-9B & 4.6 / 1.4 & 3.4 / 0.7 & 128.6 & 7.3 & 10.9 & 26.6 \\
Qwen-3.6-27B & 4.4 / 1.5 & 3.6 / 0.8 & 118.9 & 6.5 & 17.9 & 47.7 \\
Gemma-4-31B & 4.8 / 1.3 & 3.2 / 0.8 & 45.1 & 6.3 & 13.7 & 23.7 \\
Gemini-3.5-Flash & 2.3 / 1.2 & 5.7 / 1.1 & 81.2 & 8.9 & 7.0 & 18.1 \\
\midrule
\multicolumn{7}{l}{\textbf{\textit{By Benchmark (All-Model Average)}}} \\
LiveMath & 1.9 / 1.1 & 6.1 / 1.9 & 84.6 & 4.4 & 12.1 & 31.7 \\
SealQA & 4.9 / 0.9 & 3.1 / 0.4 & 98.5 & 9.4 & 15.9 & 26.9 \\
SpreadSheet & 4.5 / 1.4 & 3.5 / 1.1 & 142.5 & 9.8 & 11.3 & 38.5 \\
OfficeQA & 4.7 / 1.8 & 3.3 / 0.3 & 102.9 & 8.3 & 14.9 & 31.4 \\
ALFWorld & 3.9 / 1.6 &  4.1 / 0.8 & 93.5 & 5.8 & 14.3 & 40.6 \\
\bottomrule
\end{tabular}
\end{adjustbox}
\caption{
\textbf{Statistics of evolved skills and wiki patterns across inference models (top) and benchmarks (bottom).} For skills, we report the numbers of proposed/accepted creations and edits, along with average length in markdown lines. For wiki patterns, we report the numbers of creations and edits, along with average length in markdown lines. All wiki pattern creations and edits are retained.
}
\label{tab:skill_pattern_summary}
\vspace{-5pt}
\end{table}

\section{Related Work}
\paragraph{Experience-Driven Agent Skill Evolution}

Agent skills encode reusable procedural knowledge that allows LLM agents to leverage past experience for future tasks~\citep{zhang2025equipping, li2026skillsbench,anthropic2026buildingskills, xia2026metaclawjusttalk, zhou2026mementoskillsletagentsdesign, wang2026skillgradoptimizingagentskills, xu2026agentskillslargelanguage}. Recent frameworks enable agents to self-improve by discovering and refining procedural knowledge from past execution traces~\citep{yuksekgonul2025optimizing, agrawal2026gepa,ouyang2026skillos, xia2026skillrlevolvingagentsrecursive, lu2026skill0incontextagenticreinforcement}. Methods like EvoSkill~\citep{alzubi2026evoskill}, Trace2Skill~\citep{ni2026trace2skill}, and SkillOpt~\citep{yang2026skillopt} use specialized agent pipelines to analyze task rollouts and update modular skill documents \citep{zhang2026coevoskillsselfevolvingagentskills}. However, these methods do not maintain what has been learned as a separate, evolving knowledge representation. \method introduces a persistent Wiki Layer that consolidates experience into structured knowledge across iterations, allowing subsequent skill updates to build systematically on accumulated knowledge.

\vspace{-5pt}
\paragraph{Skill-Augmented Agents and Agent Self-Improvement}

Beyond constructing high-quality skills, skill-augmented agents must effectively select and utilize relevant skills during execution \citep{chen2026skillcraftllmagentslearn,liu2026agenticskillsworkwild}. As the number of reusable skills grows, recent work has explored skill retrieval to select relevant skills from a library for each task~\citep{zheng2026skillrouter,su2026skill,cho2026skillret,shi2026skill1,ye2026meta}. \method instead focuses on skill quality itself, separately from skill retrieval. Another line of work optimizes the broader agent harness, including prompts, context, tools, memory, and workflows~\citep{lou2026autoharness,lee2026meta,zhang2026selfharnessharnessesimprove,chen2026harnessx, lin2026agenticharnessengineeringobservabilitydriven}. These methods improve the agent system by analyzing execution traces and environment feedback to search for better agent configurations. This direction is complementary to \method, which focuses specifically on evolving reusable procedural skills while holding the broader agent harness fixed.

\vspace{-5pt}

\section{Conclusion}

We presented \method, a framework that co-evolves agent skills with a persistent, compounding knowledge base (wiki). By structuring the agent workspace into three distinct layers, \method enables skill development to build on increasingly well-supported and integrated knowledge across iterations. An orchestrated loop consolidates experience into the wiki, proposes skill refinements from accumulated knowledge, and gates changes based on validation performance. Empirically, \method consistently outperforms existing skill-evolution methods across five benchmarks and five inference models and improves over no-skill baselines in most model-dataset pairs. Beyond these overall gains, skill evolution complements model scaling: larger models generally benefit more from evolved skills, while smaller models with skills can outperform substantially larger models without them. At the same time, evolved skills transfer effectively across models and model families and can outperform self-evolved skills. Finally, our ablations confirm that persistent knowledge accumulation is critical for effective skill evolution.

\vspace{-5pt}
\section*{Limitations}

\method has several limitations that motivate future work. First, to isolate skill quality and avoid confounding effects from skill retrieval, our study follows prior work by directly injecting active skills into the agent prompt. This setup does not evaluate skill retrieval or triggering, which becomes important as the number of available skills grows. Second, our validation gating requires each accepted proposal to improve the validation score, which excludes neutral proposals that preserve immediate performance but could enable gains in subsequent iterations. We adopt this strict criterion following prior skill-evolution frameworks~\citep{yang2026skillopt,alzubi2026evoskill} to ensure a fair comparison. Exploring more flexible acceptance criteria is an important direction for future work. Third, the Wiki Layer continuously accumulates pattern pages, evolution logs, and proposal diffs across iterations, but WikiSkill currently lacks an automated mechanism to prune the wiki. Such pruning may become necessary as knowledge accumulates over longer evolution runs. Finally, while our benchmark suite includes long-context document reasoning (OfficeQA) and multi-step tool interactions, it does not cover very long-horizon tasks that span hundreds of environment actions or multiple hours. Developing online skill adaptation methods that refine procedural knowledge within a single long execution rollout remains an important direction for future work.

\vspace{-5pt}
\section*{AI Disclosure}

Large language models and coding agents are used to aid with and polish writing and generate some tables and plots.

\bibliography{main}

\clearpage
\newpage
\appendix

\section{Method Details} \label{sec:method_detail}

\subsection{Algorithm}

The full skill-evolution algorithm for \method is described in Algorithm~\ref{alg:wikiskill_evolution}.

\begin{algorithm}
\begin{algorithmic}[1]
\Require Training tasks $\mathcal{D}_{\text{train}}$, validation tasks $\mathcal{D}_{\text{val}}$, performance metric $\mathcal{R}$, iterations $K$
\State Initialize skill set $S_0 \leftarrow \emptyset$, wiki $W_0 \leftarrow \emptyset$
\State \textbf{Baseline Validation:} $\mathcal{T}_{\text{val}, 0} \leftarrow \{ \tau_i \sim \pi(x_i; S_0) \}_{x_i \in \mathcal{D}_{\text{val}}}$, \quad $\mathcal{R}_{\text{best}} \leftarrow \mathcal{R}(\mathcal{T}_{\text{val}, 0})$
\For{$k = 1, \dots, K$}
  \If{$\mathcal{R}_{\text{best}} = 1.0$}
    \State \textbf{break}
  \EndIf
  \State \textbf{Inference:} Roll out $\mathcal{T}_{\text{train}, k} \leftarrow \{ \tau_i \sim \pi(x_i; S_{k-1}) \}_{x_i \in \mathcal{D}_{\text{train}}}$
  \State Sample subset $\mathcal{T}_{\text{sample}, k} \subset \mathcal{T}_{\text{train}, k}$
  \State \textbf{Wiki Maintenance:} $W'_k \leftarrow \mathcal{M}_{\text{WM}}(W_{k-1}, \mathcal{T}_{\text{sample}, k})$
  \State \textbf{Skill Proposal:} $P_k \leftarrow \mathcal{M}_{\text{P}}(W'_k, S_{k-1}, \mathcal{T}_{\text{train}, k})$
  \State \textbf{Apply:} $S'_k \leftarrow \text{Apply}(S_{k-1}, P_k)$
  \State \textbf{Validate:} $\mathcal{T}_{\text{val}, k} \leftarrow \{ \tau_i \sim \pi(x_i; S'_k) \}_{x_i \in \mathcal{D}_{\text{val}}}$
  \If{$\mathcal{R}(\mathcal{T}_{\text{val}, k}) > \mathcal{R}_{\text{best}} $}
    \State $S_k \leftarrow S'_k$, \quad $\mathcal{R}_{\text{best}} \leftarrow \mathcal{R}(\mathcal{T}_{\text{val}, k})$, \quad $a_k \leftarrow \text{Accepted}$
  \Else
    \State $S_k \leftarrow S_{k-1}$, \quad $a_k \leftarrow \text{Rejected}$ \Comment{Roll back skills only; wiki retained}
  \EndIf
  \State \textbf{Update Wiki Log:} $W_k \leftarrow \text{Update}(W'_k, P_k, \mathcal{R}(\mathcal{T}_{\text{val}, k}), a_k)$
\EndFor
\State \Return $S_K$, $W_K$
\end{algorithmic}
\caption{\textbf{WikiSkill evolution loop.} At each iteration $k$, the inference agent rolls out on training tasks using active skills $S_{k-1}$, the maintainer consolidates sampled traces into the intermediate wiki $W'_k$, the proposer generates candidate skill modifications $P_k$, validation gating determines whether to accept $S'_k$ or roll back to $S_{k-1}$, and the system appends the proposal outcome and skill diff to produce the final wiki state $W_k$.}
\label{alg:wikiskill_evolution}
\end{algorithm}

\subsection{Distribution of Accepted Skill Updates}

We show when the updated skill proposals are accepted in Table~\ref{tab:temporal_timeline}.

\section{Dataset Details and Splits}
\label{appendix:datasets}

We describe the five benchmarks used in our evaluation below.

\textbf{LiveMathematicianBench (LiveMath)}~\citep{he2026livemathematicianbench} consists of multiple-choice mathematics competition problems from recent months. It tests the model's capacity for complex mathematical reasoning, quantifiers, and extremal conditions. \textbf{SealQA}~\citep{pham2026sealqa} is a factual question-answering benchmark composed of scholarly questions across various topics. It evaluates the agent's ability to formulate effective search queries and extract answers from web search results using a search tool. \textbf{SpreadsheetBench (SpreadSheet)}~\citep{ma2024spreadsheetbench} tests the agent's ability to write correct code under library constraints (such as formula evaluation limitations) and execute complex table transformations. \textbf{OfficeQA}~\citep{singhvi2025officeqa} evaluates long-context question-answering over a large repository of historical Treasury bulletins. Tasks require synthesizing evidence across long contexts and multi-page financial tables. Following the setup in \citet{yang2026skillopt}, the agent is provided with pre-parsed oracle reference pages as initial document evidence in the prompt, while retaining access to local text-processing tools (\texttt{glob}, \texttt{grep}, \texttt{read}) to search, cross-reference, and inspect full Treasury bulletin files on disk. \textbf{ALFWorld}~\citep{shridhar2021alfworld} is an interactive text-based embodied environment where an agent solves multi-step household tasks (e.g., picking and placing objects, heating or cooling items) by outputting text actions to a simulator. Unlike static QA benchmarks, ALFWorld tests sequential decision-making, spatial reasoning, and error recovery from simulator feedback.

\begin{table}[t]
\centering
\small
\begin{tabular}{lccc}
\toprule
\textbf{Category} & \textbf{Early (Iter 0--1)} & \textbf{Mid (Iter 2--4)} & \textbf{Late (Iter 5--7)} \\
\midrule
\multicolumn{4}{l}{\textbf{\textit{By Model}}} \\
Qwen-3.5-4B  & 39\% & 39\% & 21\% \\
Qwen-3.5-9B  & 52\% & 30\% & 19\% \\
Qwen-3.6-27B  & 43\% & 40\% & 17\% \\
Gemma-4-31B  & 52\% & 37\% & 11\% \\
Gemini-3.5-Flash & 50\% & 46\% & 4\% \\
\midrule
\multicolumn{4}{l}{\textbf{\textit{By Benchmark}}} \\
LiveMath & 44\% & 42\% & 14\% \\
SealQA & 39\% & 33\% & 28\% \\
SpreadSheet & 41\% & 48\% & 11\% \\
OfficeQA & 58\% & 26\% & 16\% \\
ALFWorld & 55\% & 34\% & 10\% \\
\bottomrule
\end{tabular}
\caption{
\textbf{Distribution of accepted skill updates across evolution iterations, grouped by model (top) and benchmark (bottom).} Percentages indicate the proportion of accepted updates that occur during each stage of evolution.
}
\label{tab:temporal_timeline}
\vspace{-5pt}
\end{table}

\begin{table}[h]
\centering
\small
\begin{tabular}{lccccc}
\toprule
\textbf{Benchmark} & \textbf{Interaction} & \textbf{Train} & \textbf{Val} & \textbf{Test}  & \textbf{Environment Tools} \\
\midrule
LiveMath & Single-Step & 35 & 18 & 124 & None (Direct Reasoning) \\
SealQA & Multi-Step & 16 & 10 & 85 & \texttt{web\_search}, \texttt{read\_file} \\
SpreadSheet & Multi-Step & 80 & 40 & 280 & \texttt{bash} \\
OfficeQA & Multi-Step & 50 & 24 & 172 & \texttt{glob}, \texttt{grep}, \texttt{read} \\
ALFWorld & Multi-Step & 39 & 18 & 134 & Admissible Actions \\
\bottomrule
\end{tabular}
\caption{
\textbf{Benchmark statistics, data splits, interaction modes, and available environment tools.}
}
\label{tab:dataset_splits}
\end{table}

Table~\ref{tab:dataset_splits} summarizes the sample counts across training, validation, and test splits, interaction modes, and available tools for each benchmark evaluated in our experiments. All task splits and available toolsets are strictly matched with prior work \citep{yang2026skillopt, alzubi2026evoskill}. For tool setup, LiveMath operates as a single-step reasoning benchmark without external tools, where the model generates final answers directly; SealQA equips the agent with web search (using Google Search API) and file reading for multi-step factual retrieval (we use the July, 2026 version of SealQA for all experiments); SpreadSheet provides a \texttt{bash} shell tool for Python code execution and table manipulation; OfficeQA provides local text-search utilities for multi-step Treasury bulletin navigation; and ALFWorld provides an interactive simulator action space for multi-step embodied decision making.

\paragraph{Evaluation robustness with small validation sets}
Following established setups in prior work~\citep{yang2026skillopt, alzubi2026evoskill}, benchmark validation splits are relatively small, which can introduce evaluation noise into gating decision. To account for this variability, all reported scores represent the average test performance across three independent runs of the entire evolutionary pipeline, with paired bootstrap significance testing (detailed in Appendix~\ref{appendix:implementation}).

\section{Implementation Details}
\label{appendix:implementation}

To provide diagnostic feedback for the Wiki maintainer, we apply a stratified sampling strategy ($\mathcal{T}_{\text{sample}, k} \subset \mathcal{T}_{\text{train}, k}$) at each iteration $k$. Specifically, the system samples up to 8 traces per iteration, stratified into a maximum of 5 failing traces (to perform root-cause analysis of errors) and up to 3 passing traces (to identify effective strategies and prevent regressions in working behaviors). Each individual execution log is capped at 15,000 characters prior to injection into the prompt.

\paragraph{Statistical significance testing} We perform paired bootstrap significance tests with $1,000$ iterations for each benchmark. In each bootstrap iteration, task instances are sampled with replacement from the test split $\mathcal{D}_{\text{test}}$ to construct a bootstrap evaluation set of size $|\mathcal{D}_{\text{test}}|$, from which candidate accuracy scores and pairwise performance margins are computed. To evaluate overall cross-benchmark performance, we conduct stratified macro-average bootstrap resampling: in each iteration, task instances are resampled independently with replacement within each benchmark, and we calculate macro-average accuracy by assigning equal weight to all benchmarks.

We determine top-performing methods in our evaluation tables as follows. Methods are initially ranked by their observed performance (or macro-average performance across benchmarks). A top-ranked method $M^*$ is the sole top performer if and only if it achieves a statistically significant gain over all competing methods at $p < 0.05$. If $M^*$ is not statistically distinguishable ($p \ge 0.05$) against one or more lower-ranked methods, no single method is declared the sole top performer. Instead, all methods whose performance is not statistically distinguishable from $M^*$ ($p \ge 0.05$) are grouped into a top-tier statistical tie (and bolded accordingly).

\section{Baseline Details and Optimizer API Call Analysis}
\label{appendix:api_complexity}

\subsection{Baseline Methods}

\paragraph{Trace2Skill~\citep{ni2026trace2skill}}
Trace2Skill employs a three-stage pipeline centered on parallel trace analysis and hierarchical merging. It evaluates the current skill on training tasks and dispatches parallel success analysts and error analysts to extract effective strategies from passing tasks and diagnose root causes of failures. The resulting structured patches are recursively consolidated via a hierarchical merge operator into a single patch set. The consolidated patches are applied to the skill document and accepted based on validation performance.

\paragraph{EvoSkill~\citep{alzubi2026evoskill}}
EvoSkill frames skill evolution as a search over a frontier of candidate programs. In each iteration, it samples training tasks via a round-robin category schedule and executes rollouts. EvoSkill feeds \textit{only} failure traces to the proposer alongside a flat feedback history of past proposal outcomes. The proposer generates candidate skill modifications that are materialized into \texttt{SKILL.md} files, scored on a validation split, and added to a bounded frontier of top-performing programs.

\paragraph{SkillOpt~\citep{yang2026skillopt}}
SkillOpt implements a six-stage \textit{ReflACT} pipeline (Rollout, Reflect, Aggregate, Select, Update, Evaluate) for iterative skill optimization. In each epoch, the system rolls out the agent on training tasks and reflects on full execution traces, including both successes and failures, to generate candidate patches. These patches are hierarchically aggregated and selected to update a single monolithic skill document, which is accepted or rejected based on validation performance.

\subsection{Optimizer API Call Complexity}

In this section, we analyze optimizer API call complexity, denoted by $\mathcal{C}$, across self-improving agent frameworks. We define one evolution iteration as rolling out the agent on the full training split $\mathcal{D}_{\text{train}}$ once ($N_{\text{train}} = |\mathcal{D}_{\text{train}}|$ training task instances), processed in minibatches of size $B$ ($B \le N_{\text{train}}$). Table~\ref{tab:api_complexity_summary} summarizes the per-iteration optimizer API call complexity for WikiSkill and prior methods.
\begin{table}[h]
\centering
\small
\begin{tabular}{l c c}
\toprule
\textbf{Framework} & \textbf{Per-Iteration Formula} & \textbf{Complexity} \\
\midrule
Trace2Skill & $N_{\text{train}} + \left(1 + \frac{1}{c-1}\right) \frac{N_{\text{train}}}{B} + 1$ & $\mathcal{O}\left(N_{\text{train}} + \frac{N_{\text{train}}}{B}\right)$ \\
EvoSkill & $\frac{2 N_{\text{train}}}{B}$ & $\mathcal{O}\left(\frac{N_{\text{train}}}{B}\right)$ \\
SkillOpt & $\frac{K_{\text{opt}} \cdot N_{\text{train}}}{B}$ & $\mathcal{O}\left(\frac{N_{\text{train}}}{B}\right)$ \\
\methodcolor{\textbf{WikiSkill}} & \methodcolor{$(1 + T_{\text{ReAct}}) \frac{N_{\text{train}}}{B}$} & \methodcolor{$\mathcal{O}\left(\frac{N_{\text{train}}}{B}\right)$} \\
\bottomrule
\end{tabular}
\caption{\textbf{Comparison of optimizer API call complexity per evolution iteration across self-improving agent frameworks.} $N_{\text{train}} = |\mathcal{D}_{\text{train}}|$ denotes the number of training tasks, $B$ denotes the batch size, $T_{\text{ReAct}}$ denotes the number of interactive ReAct reasoning turns used by the Skill Proposer agent in WikiSkill, $K_{\text{opt}}$ denotes the number of reflection and merging calls per step in SkillOpt, and $c$ denotes the reduction-tree branching factor in Trace2Skill. \method uses $B = N_{\text{train}}$ across all datasets.
In full-batch mode, \method's optimizer call count is independent of training set size $N_{\text{train}}$, requiring $1 + T_{\text{ReAct}}$ optimizer calls per iteration.}

\label{tab:api_complexity_summary}
\end{table}

\paragraph{WikiSkill}
For each batch of size $B$, the Wiki Maintainer requires one LLM call to analyze sampled traces $\mathcal{T}_{\text{sample}, k}$ and consolidate pattern pages into the intermediate wiki $W'_k$. The Skill Proposer then runs as an autonomous multi-turn ReAct agent, executing interactive tool calls over $T_{\text{ReAct}}$ reasoning turns (roughly $10 \le T_{\text{ReAct}} \le 20$ across our experiment runs), where each ReAct turn requires $1$ LLM call. When processing training data in batches of size $B$, completing one full iteration over $N_{\text{train}}$ tasks requires $\frac{N_{\text{train}}}{B}$ steps:
\begin{equation}
\mathcal{C}_{\text{WikiSkill}} = (1 + T_{\text{ReAct}}) \frac{N_{\text{train}}}{B}
\end{equation}
In our experiments, we set the batch size to the full training size ($B = N_{\text{train}}$) across all datasets. In this full-batch setting ($\frac{N_{\text{train}}}{B} = 1$), $\mathcal{C}_{\text{WikiSkill}} = 1 + T_{\text{ReAct}}$. Because $T_{\text{ReAct}}$ does not depend on $N_{\text{train}}$, \method's optimizer API call complexity is $\mathcal{O}(1)$ with respect to training set size. Specifically, each iteration requires $1 + T_{\text{ReAct}}$ optimizer LLM calls, regardless of the number of training instances. While this constant call complexity may incur higher inference cost on some datasets, the additional computation is accompanied by consistent performance gains over prior skill-evolution methods across our evaluation.

\paragraph{EvoSkill}
EvoSkill partitions training tasks into minibatches of size $B$. For each minibatch, EvoSkill uses one Proposer LLM call for error diagnosis and one Generator LLM call for skill updates. Processing the full training split of $N_{\text{train}}$ tasks requires $\frac{N_{\text{train}}}{B}$ minibatch steps, resulting in:
\begin{equation}
\mathcal{C}_{\text{EvoSkill}} = \frac{2 N_{\text{train}}}{B}
\end{equation}
Thus, EvoSkill's optimizer API call complexity scales linearly with training set size $N_{\text{train}}$ ($\mathcal{O}(N_{\text{train}}/B)$).
\paragraph{SkillOpt}
SkillOpt evaluates minibatches of size $B$, completing each iteration in $\frac{N_{\text{train}}}{B}$ optimization steps. During each step, SkillOpt executes its ReflACT pipeline (parallel analyst reflections, hierarchical patch synthesis, and candidate selection), requiring $K_{\text{opt}} \approx 6\text{--}8$ optimizer LLM calls per step:
\begin{equation}
\mathcal{C}_{\text{SkillOpt}} = \frac{K_{\text{opt}} \cdot N_{\text{train}}}{B}
\end{equation}
SkillOpt similarly scales linearly with training set size $N_{\text{train}}$, as $\mathcal{O}(N_{\text{train}}/B)$.
\paragraph{Trace2Skill}
Trace2Skill processes training trajectories through a three-stage pipeline per iteration:
\begin{enumerate}[wide=0pt]
\item \textbf{Trace analysis stage:} Every individual execution trajectory is analyzed independently with one LLM call, incurring $N_{\text{train}}$ total calls.
\item \textbf{Patch map stage:} Analysis records are chunked into batches of size $B$, requiring $\frac{N_{\text{train}}}{B}$ calls to generate local skill patches.
\item \textbf{Hierarchical reduce \& apply stage:} The $\frac{N_{\text{train}}}{B}$ local patches are recursively merged via a $c$-ary reduction tree (where $c$ is the branching factor). Summing across levels yields $\approx \frac{1}{c-1} \frac{N_{\text{train}}}{B}$ merge calls, plus one final call to format the skill document.
\end{enumerate}
Combining all stages:
\begin{equation}
\mathcal{C}_{\text{Trace2Skill}} \approx N_{\text{train}} + \left(1 + \frac{1}{c-1}\right) \frac{N_{\text{train}}}{B} + 1
\end{equation}
Because Trace2Skill performs individual LLM analysis on every training trajectory ($N_{\text{train}}$ calls), its complexity is lower-bounded by $\mathcal{O}(N_{\text{train}})$, scaling linearly with training set size.

\paragraph{Full-batch training vs. minibatch optimization} Across all datasets, we set the batch size $B$ to the full training set size ($B = N_{\text{train}}$) for WikiSkill, processing the entire training set at once per iteration. The Skill Proposer dynamically searches, selects, and reads specific execution traces on demand to diagnose root causes before proposing skill updates. In contrast, EvoSkill and SkillOpt achieve their best performance under minibatch settings ($B < N_{\text{train}}$), which causes their optimizer API call complexity to scale linearly with training set size. Finally, Trace2Skill remains strictly $\mathcal{O}(N_{\text{train}})$ regardless of minibatch size because it requires an independent LLM call for every training trajectory.

\section{System and Agent Prompts}
\label{appendix:prompts}
We provide the exact system prompts used for (1) the Inference Agent for each task across all methods, (2) the Wiki Maintainer, and (3) the Skill Proposer in WikiSkill.

\subsection{Task Inference Agent System Prompts}
\label{appendix:prompts:task_agents}
\filehead{skillhead}{LiveMathematicianBench Inference Agent System Prompt}
\begin{lstlisting}[style=md]
You are an expert mathematical reasoning agent solving multiple-choice questions.

{skill_section}

## Task Format
You will receive one mathematics multiple-choice question and its answer choices. Reason carefully about quantifiers, hypotheses, extremal wording, and exact equality conditions.

## Answer Format
Think step by step, then provide your final answer inside <answer>...</answer> tags. Inside the tags, output only the single choice label, such as A or C.

Example:
<answer>B</answer>
\end{lstlisting}
\vspace{1em}
\filehead{skillhead}{SealQA Inference Agent System Prompt}
\begin{lstlisting}[style=md]
You are a knowledgeable question-answering assistant with access to web_search and read_file tools.

{skill_section}

## Task
You will receive a factual question. To answer it:
1. You can check the available skills. They contain guidance that can improve your search queries and answer accuracy.
2. You can use web_search to find relevant information. You can call it multiple times with different queries.
3. You can do web_search anytime during the process depending on your needs.
4. After gathering enough information, provide your final answer.

## Answer Format
You MUST wrap your final answer in <answer> tags:
<answer>
... your final answer (exact value only, no explanation) ...
</answer>
\end{lstlisting}
\vspace{1em}
\filehead{skillhead}{SpreadsheetBench Inference Agent System Prompt}
\begin{lstlisting}[style=md]
You are a spreadsheet expert who can manipulate spreadsheets through Python code.

{skill_section}

You need to solve the given spreadsheet manipulation question, which contains the following information:
- working_directory: The absolute path to your working directory where files are located.
- instruction: The question about spreadsheet manipulation.
- spreadsheet_path: The absolute path of the spreadsheet file you need to manipulate.
- spreadsheet_content: The first few rows of the content of spreadsheet file.
- instruction_type: There are two values (Cell-Level Manipulation, Sheet-Level Manipulation) used to indicate whether the answer to this question applies only to specific cells or to the entire worksheet.
- answer_position: The position need to be modified or filled. For Cell-Level Manipulation questions, this field is filled with the cell position; for Sheet-Level Manipulation, it is the maximum range of cells you need to modify. You only need to modify or fill in values within the cell range specified by answer_position.
- output_path: The absolute path where you must save the modified spreadsheet.

## CRITICAL RESTRICTIONS

You can ONLY read and write files within the **working_directory**. Any attempt to access files outside this directory will fail.

- **Allowed paths**: working_directory (and its subdirectories)
- **Read from**: spreadsheet_path (inside working_directory)
- **Write to**: output_path (inside working_directory)

Do NOT create files outside the working_directory. Use the exact absolute paths provided.

You have access to a bash tool that can execute any shell command.
\end{lstlisting}
\vspace{1em}
\filehead{skillhead}{OfficeQA Inference Agent System Prompt}
\begin{lstlisting}[style=md]
You are an expert OfficeQA agent working over local Treasury bulletin text files.

{skill_section}

## Rules
1. Use only the provided local document tools to inspect candidate files.
2. Narrow to the most relevant file before reading long passages.
3. Prefer short targeted searches, then small reads around matching evidence.
4. Do not invent values that are not grounded in the retrieved text.
5. When the question requires arithmetic, compute only after extracting the exact operands.
6. If you have enough evidence, return the final answer inside <answer>...</answer>.

## Tool Use
Use the provided function tools directly when you need them. Prefer searching and small reads before answering. Do not ask the user for permission to use tools; just call the tools.

## Final Answer Format
When you are ready to answer, emit the final answer inside <answer>...</answer> and do not request another tool.
\end{lstlisting}
\vspace{1em}
\filehead{skillhead}{ALFWorld Inference Agent System Prompt}
\begin{lstlisting}[style=md]
You are an expert agent operating in the ALFRED Embodied Environment. Your task is to: {task_description}

{skill_section}

Prior to this step, you have already taken {step_count} step(s). Below are the most recent {history_length} observations and the corresponding actions you took: {action_history}

You are now at step {current_step} and your current observation is: {current_observation}

Your admissible actions of the current situation are: [{admissible_actions}].

Now it's your turn to take an action. You should first reason step-by-step about the current situation. This reasoning process MUST be enclosed within <think> </think> tags. Once you've finished your reasoning, you should choose an admissible action for current step and present it within <action> </action> tags.
\end{lstlisting}
\vspace{1.5em}
\subsection{Wiki Maintainer Agent System Prompt}
\label{appendix:prompts:wiki_maintainer}
\filehead{skillhead}{Wiki Maintainer Agent System Prompt}
\begin{lstlisting}[style=md]
You are a Wiki Maintainer Agent for an LLM skill evolution system.

Your job is to maintain a structured knowledge base (wiki) that documents patterns observed during agent execution -- both successes and failures. You must perform DEEP ANALYSIS of execution logs to identify root causes, not just surface-level symptoms.

## Wiki Structure
The wiki is organized as:
- wiki/index.md -- Concise catalog of known patterns (one line per pattern)
- wiki/log.md -- Chronological evolution log (iterations, scores, accept/reject)
- wiki/skill-impact.md -- Record of which skills were tried and their outcomes
- wiki/patterns/ -- One page per pattern with detailed evidence and analysis

## Your Input
1. Execution traces from the latest iteration -- including full agent execution logs showing what actions the agent took, what commands it ran, and what environment feedback it observed
2. The current wiki context (index, log, pattern pages)

## Your Output (Incremental Edit Mode)
Return a JSON object with these keys:
- "create_patterns": list of {"name": "pattern-name.md", "content": "..."} -- new patterns (full content)
- "update_patterns": list of {"name": "existing-pattern.md", "edits": [...]} -- patch existing patterns
- "update_index": full updated content of index.md (always provide the complete index)
- "append_log": "brief summary of this iteration's findings and actions"

"update_index" and "append_log" are REQUIRED. Always provide them, even if there are no new patterns. For "update_index", always provide the complete updated index content including all existing entries plus any new ones.

### Patch Operations (for update_patterns only)
For "update_patterns", each entry uses an "edits" list of patch operations:
- {"op": "append", "content": "text to add at end"}
- {"op": "replace", "target": "exact text to find", "content": "replacement text"}
- {"op": "insert_after", "target": "exact text to find", "content": "text to insert after"}

Rules for patch operations:
1. "target" must be an EXACT substring of the existing content.
2. Use "append" to add new evidence. Use "replace" to fix or refine existing text.
3. Use "insert_after" to add entries after a specific line.
4. Keep each edit minimal -- only change what's needed.
5. For NEW patterns (create_patterns), use full "content".

## Analysis Guidelines

### Deep Trace Analysis (CRITICAL)
When execution logs are provided, you MUST:
1. Read the agent's actual actions -- what commands did it issue?
2. Compare successful vs failed tasks -- what did successful tasks do differently?
3. Identify ACTION PATTERNS and strategies, not just error messages.
4. Check whether the agent followed any active skills, and whether the skill guidance was helpful or not
   
### Pattern Documentation Rules
1. Each pattern page should document:
   - What the pattern is (description)
   - Root cause analysis (WHY it happens, not just WHAT happens)
   - Exact command sequences from traces (what the agent did wrong / right)
   - Known solutions or workarounds (concrete action patterns with exact syntax)
2. Capture BOTH success and failure patterns:
   - **Failure patterns**: Document what went wrong and how to avoid it
   - **Success patterns**: Document strategies that consistently lead to task completion
3. Do NOT create duplicate patterns -- update existing ones with new evidence
4. Be concise. Pattern pages should be 10-30 lines, not essays.
5. Only create patterns for meaningful, generalizable observations.

### Index Description Quality (CRITICAL)
The index.md entries are the MOST IMPORTANT part of the wiki because they determine whether inference agents will read the full pattern pages.

Each index entry MUST follow this format:
- [pattern-name](wiki/patterns/pattern-name.md): PROBLEM + ROOT CAUSE + FIX in one or two sentence.

The description must be specific enough that an agent can judge relevance without reading the full page. Include the problem, root cause, AND solution.
\end{lstlisting}
\vspace{1.5em}
\subsection{Skill Proposer Agent System Prompt (ReAct Mode)}
\label{appendix:prompts:proposer}

\filehead{skillhead}{Skill Proposer Agent System Prompt}
\begin{lstlisting}[style=md]
You are a Skill Proposer Agent for an LLM agent that solves {task_desc}.
Your job is to explore the wiki knowledge base and execution traces, diagnose root causes of failures, and propose a skill change (create or patch).

## Tools Available
You have two tools:
1. `read_file(path)` -- Read a wiki file or execution log. Paths are relative to the workspace root.
2. `finish(proposal)` -- Submit your final skill proposal as a JSON object.

## Workflow
1. Start by reading `wiki/index.md` to understand what patterns exist
2. Read `wiki/skill-impact.md` to see what was tried before (includes full content of rejected proposals -- DO NOT repeat rejected approaches)
3. Read specific pattern pages that seem relevant to the current failures
4. Read execution traces for failed tasks via `traces/<task_id>` to understand root causes
5. Decide: create (new skill) or patch (edit existing skill), or no_action
6. If proposing a change, call `finish` with the full proposal

## finish() Proposal Format
For creating a new skill:
- "action": "create"
- "name": skill directory name (snake_case)
- "skill_md": full SKILL.md content with YAML frontmatter + When to Apply + When NOT to Apply + Instructions
- "purpose_md": full PURPOSE.md content with Origin + Patterns Addressed + Evolution History

For patching an existing skill:
- "action": "patch"
- "name": existing skill directory name
- "edits": list of patch operations:
  - {"op": "append", "content": "text to add at end"}
  - {"op": "replace", "target": "exact text to find", "content": "replacement"}
  - {"op": "insert_after", "target": "exact text to find", "content": "text to insert after"}
  Each "replace" target should be a short, specific section -- not the entire file. If you need to change most of the file, use "action": "create" instead.

If no action is needed, call finish with: {"action": "no_action"}

## Rules
1. Read the wiki FIRST -- don't propose something that was already tried and rejected. skill-impact.md contains full content of rejected proposals.
2. Focus on action patterns and concrete strategies.
3. Keep skills concise and actionable.
4. You MUST read at least 4 execution traces before proposing a skill change. Target your exploration based on the trace summary.
5. Prefer patching existing skills over creating new ones when the existing skill is partially correct.
\end{lstlisting}

Note that execution traces are physically stored in the Raw Layer (\texttt{raw/traces/}), while the workspace environment resolves \texttt{read\_file("traces/<task\_id>")} calls by automatically mapping the \texttt{traces/} alias to the corresponding execution log under \texttt{raw/} for the Skill Proposer.

\end{document}